\pdfoutput=1

\documentclass[11pt]{article}

\usepackage[]{ACL2023}

\usepackage{times}
\usepackage{latexsym}

\usepackage[T1]{fontenc}

\usepackage[utf8]{inputenc}

\usepackage{microtype}

\usepackage{inconsolata}
\usepackage{booktabs}
\usepackage{graphicx}
\usepackage{soul}
\usepackage[dvipsnames]{xcolor}
\usepackage[most]{tcolorbox}
\usepackage{tabularx}
\usepackage{enumitem}
\usepackage{float}
\usepackage{bbding}
\usepackage{pifont}

\title{Integrating Video and Text: A Balanced Approach to Multimodal Summary Generation and Evaluation}

\author{
\textbf{
Galann Pennec\textsuperscript{$\infty,\diamondsuit,\heartsuit$} 
\ \quad
Zhengyuan Liu\textsuperscript{$\diamondsuit,\heartsuit$} 
\ \quad 
}
\\
\textbf{
Nicholas Asher\textsuperscript{$\S,\heartsuit$} 
\ \quad 
Philippe Muller\textsuperscript{$\infty,\heartsuit$} 
\ \quad 
Nancy F. Chen\textsuperscript{$\diamondsuit,\heartsuit$}
}
\\
\textsuperscript{$\infty$}IRIT, University of Toulouse, France\\
\textsuperscript{$\diamondsuit$}Institute for Infocomm Research (I$^2$R), A*STAR, Singapore\\
\textsuperscript{$\heartsuit$}CNRS@CREATE, Singapore
\quad
\textsuperscript{$\S$}CNRS, IRIT, France\\
\texttt{galann.pennec@cnrsatcreate.sg},
\ \texttt{\{liu\_zhengyuan,nancy\_chen\}@a-star.edu.sg}\\
\ \texttt{\{nicholas.asher,philippe.muller\}@irit.fr}
}

\begin{document}

\newcommand{\cit}{{\color{red} cite }}

\newcommand{\hlred}[1]{{\sethlcolor{red}\hl{#1}}}
\newcommand{\hly}[1]{{\sethlcolor{yellow}\hl{#1}}}
\newcommand{\hlgr}[1]{{\sethlcolor{green}\hl{#1}}}
\newcommand{\boxred}[1]{{\fcolorbox{red}{white}{#1}}}
\newcommand{\ZY}[1]{\textcolor{purple}{[ZY: #1]}}
\newcommand{\DONE}[1]{\textcolor{gray}{[DONE: #1]}}

\maketitle
\begin{abstract}

Vision-Language Models (VLMs) often struggle to balance visual and textual information when summarizing complex multimodal inputs, such as entire TV show episodes. In this paper, we propose a zero-shot video-to-text summarization approach that builds a screenplay-like representation of an episode, effectively integrating key video moments, dialogue, and character information into a unified document. Unlike previous approaches, we simultaneously generate screenplays and name the characters in zero-shot, using only the audio, video, and transcripts as input.
Additionally, we highlight that existing summarization metrics can fail to assess the multimodal content in summaries.
To address this, we introduce~\textsc{MFactSum}, a multimodal metric that evaluates summaries with respect to both vision and text modalities. Using~\textsc{MFactSum}, we evaluate our screenplay summaries on the SummScreen3D dataset, demonstrating superiority against state-of-the-art VLMs such as Gemini 1.5 by generating summaries containing 20\% more relevant visual information while requiring 75\% less of the video as input.\footnote{Codes of this work are available at~\url{https://github.com/galannp/integrating_video_and_text}}

\end{abstract}

\begin{figure*}[hbt]
\begin{center}
\includegraphics[width=0.75\textwidth]{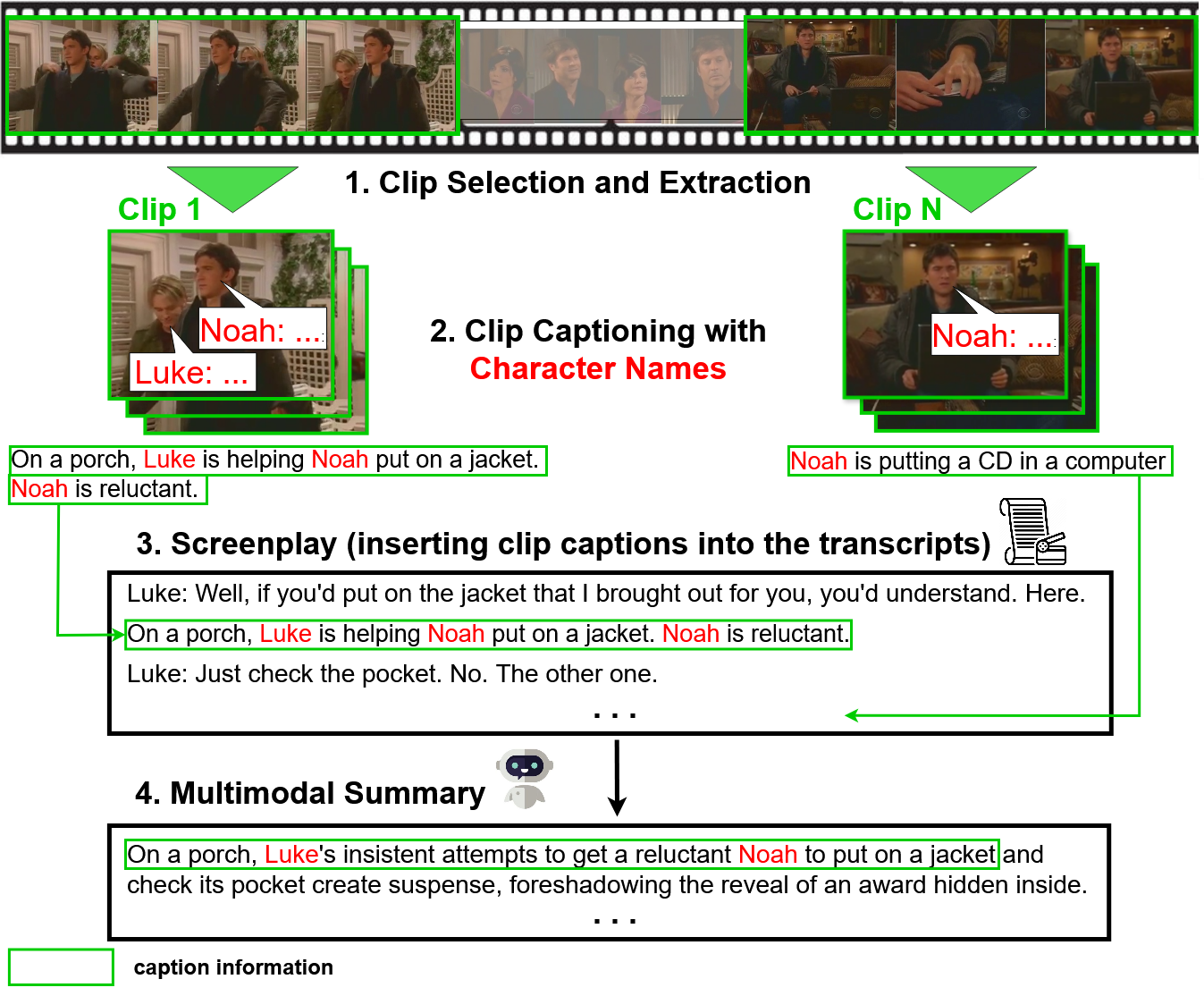}
\end{center}
\caption{\textbf{Our zero-shot pipeline for summarizing long TV show episodes} 1) We select and extract important video clips from the whole episode 2) We provide a video clip and its corresponding transcripts to a VLM for caption generation while reidentifying featured characters 3) Screenplays are built by inserting the clip captions at the correct timestamp in the full episode transcripts 4) We finally perform summarization of those screenplays}
\label{fig:pipeline}
\end{figure*}

\section{Introduction}

VLMs~\citep{DBLP:journals/corr/abs-2407-21783, DBLP:journals/corr/abs-2406-07476, GPT-4o} still struggle to effectively balance both vision and text modalities in their answers, sometimes neglecting or completely ignoring one input modality over the other~\citep{DBLP:journals/corr/abs-2403-05262, DBLP:journals/corr/abs-2401-09774, DBLP:conf/cvpr/ShenYWYEW22, DBLP:journals/corr/abs-2408-12763}. This challenge is particularly present in video-text tasks, where datasets often suffer from poor annotations and limited diversity, restricting the model's ability to bridge the gap between vision and text~\citep{DBLP:journals/corr/abs-2404-12353}.
Furthermore, there are significant uncertainties regarding how VLMs handle long multimodal contexts, such as hour-long videos or sequences of hundreds of images~\citep{DBLP:journals/corr/abs-2405-21075, DBLP:journals/corr/abs-2404-18532, DBLP:journals/corr/abs-2406-08035, DBLP:journals/corr/abs-2406-04264, DBLP:journals/corr/abs-2406-16852}.

To address these challenges, storytelling approaches have emerged as promising alternatives. These methods aim to improve multimodal understanding by first generating textual descriptions from videos (e.g., scripts, screenplays) before applying them to various downstream tasks~\citep{DBLP:journals/corr/abs-2406-17309, DBLP:conf/emnlp/0010LIWYBB24, DBLP:conf/emnlp/BhattacharyyaSK23}. However, little is known about how storytelling approaches compare to state-of-the-art VLMs in summarizing long videos and transcripts.

In this paper, we propose a zero-shot multimodal pipeline (Figure~\ref{fig:pipeline}) to summarize long videos, such as TV show episodes, by building our own textual screenplays. We argue that screenplays provide a natural way for both human users (e.g. actors) and Large Language Models (LLMs) to access and interpret all the multimodal content of an episode into a unified document, including dialogue, character interactions, emotions, behaviors and scene locations. Those screenplay representations are reusable across different tasks~(see Section~\ref{sec:related}), thereby limiting the need to reencode the whole video every time.

Unlike previous storytelling approaches, we both generate video captions and reidentify characters at the same time by prompting a VLM in zero-shot on the audio, video and transcripts. Additionally, we reduce the generation cost of our screenplays while preserving their multimodal richness. We do this by identifying and summarizing key moments from the video (Figure~\ref{fig:pipeline}), since we observe that such moments often align with natural pauses in the dialogue.

Moreover, we find that existing summarization metrics tend to overlook visual content understanding, as such information is often less present in TV show summaries.
This motivates us to introduce a new evaluation metric, \textsc{MFactSum}, designed to assess the multimodal fidelity of video-to-text summaries. Such a metric is crucial for the task, as text modality alone cannot fully capture emotions, actions, and locations that provide essential context to an episode. Inspired by factual consistency metrics such as FactScore~\citep{DBLP:conf/emnlp/MinKLLYKIZH23} and PRISMA~\citep{DBLP:conf/acl/MahonL24}, our metric evaluates the recall of both visual facts (video related) and textual facts (transcripts related) within a generated summary. By separately assessing visual and textual understanding, \textsc{MFactSum} provides a balanced evaluation of a summarization system’s multimodal capabilities.

In summary, our contributions are as follows:

\begin{itemize}
    \item We introduce a zero-shot multimodal approach (Figure~\ref{fig:pipeline}) that summarizes TV shows by building a screenplay-like document integrating key video moments, transcripts, and character information.

    \item We propose a new multimodal metric,~\textsc{MFactSum} to evaluate how well a summarization system equally captures both textual and visual information.

    \item We evaluate our pipeline on SummScreen3D\footnote{\url{https://github.com/ppapalampidi/long_video_summarization}}~\citep{DBLP:conf/eacl/PapalampidiL23} containing long soap opera episodes. Our results show that our screenplay-based summaries retrieve~20\% more visual information than Gemini 1.5 Pro while using about~75\% less of the video as input.
\end{itemize}

Through this work, we aim to improve summarization models and evaluation protocols by making them more balanced across modalities, ultimately advancing the field of multimodal summarization.

\section{Related Work} \label{sec:related}

\paragraph{Multimodal Summarization of Movies and TV Shows}
The task can be performed in two different ways depending on the nature of the summary.
While the output summary can be a video (a recap or trailer)~\citep{DBLP:conf/cvpr/SinghST24, DBLP:conf/aaai/PapalampidiKL21, DBLP:conf/cikm/ChenZZ24}, our study instead focuses on generating textual summaries from a TV show episode.
Originally, textual summaries were produced given scripts or screenplays of the movie or TV show episode~\citep{DBLP:conf/acl/SaxenaK24, DBLP:conf/naacl/GorinskiL15}.
More recently, multimodal summarization datasets like SummScreen3D~\citep{DBLP:conf/eacl/PapalampidiL23}, presented in detail in Section~\ref{subsec:datasets}, have led the way to new Vision-Language approaches for the task~\citep{DBLP:conf/eacl/PapalampidiL23, DBLP:conf/acl/MahonL24}.

Yet, little is known about how these models leverage both modalities when summarizing hour-long TV show episodes. We therefore propose~\textsc{MFactSum} as a new multimodal evaluation strategy.

\paragraph{Storytelling Methods} \label{subsec:related_storytelling}

Movie audio description~\citep{DBLP:conf/cvpr/RohrbachRTS15, DBLP:conf/cvpr/SoldanPAH0GG22} or movie screenplays~\citep{DBLP:conf/naacl/GorinskiL15, DBLP:conf/acl/SaxenaK24} fall into the same category of documents that relate the visual elements of a movie and put them back in the context of the story for a complete and accurate multimodal understanding.

When available for a movie, screenplays or other forms of narration can serve as a basis to perform any downstream task such as movie summarization~\citep{DBLP:conf/acl/SaxenaK24, DBLP:conf/naacl/GorinskiL15, DBLP:conf/emnlp/SangMYWLS22, DBLP:conf/naacl/GorinskiL18, DBLP:journals/mms/ReboudHLT23, DBLP:conf/eccv/HuangXRWL20}.

Some recent approaches known as storytelling methods even automatically generate their own screenplays or textual descriptions of a movie to later perform video understanding~\citep{DBLP:journals/corr/abs-2406-17309, DBLP:conf/emnlp/0010LIWYBB24, DBLP:conf/emnlp/BhattacharyyaSK23}, trailer prediction~\citep{DBLP:conf/cikm/ChenZZ24} or even allow the user to have a conversation over an entire movie~\citep{DBLP:journals/corr/abs-2310-19773}.
In this paper, we generate our own screenplays from the input video and transcripts and use them for summarizing entire TV show episodes (see Section~\ref{sec:summ_model}).

\paragraph{Character Identification}
Characters constitute a central part of any story. Character identification in a video helps the models better connect what is displayed on screen to the actual conversations, eventually improving multimodal understanding of movies and TV shows.

Applications of character identification in story understanding include Video Question Answering (VideoQA)~\citep{DBLP:journals/corr/abs-2005-08646, DBLP:conf/acl/LeiYBB20, DBLP:conf/aaai/ChoiOHSJLZ21}, movie audio description~\citep{DBLP:journals/mta/PiniCBBC19, DBLP:conf/iccv/HanBNVXZ23, DBLP:conf/cvpr/HanBNVXZ24, DBLP:journals/corr/abs-2403-12922, DBLP:journals/corr/abs-2407-15850, DBLP:conf/cvpr/RaajeshDKT24}, dialogue and movie summarization~\citep{liu-2019-topic,DBLP:conf/mm/SangX10, DBLP:journals/mta/TranHLJ17,liu-2021-coreference}.
The task can be performed based on the video coupled with metadata (e.g. from IMDb\footnote{\url{https://www.imdb.com}})~\citep{DBLP:conf/iccv/HanBNVXZ23, DBLP:conf/cvpr/HanBNVXZ24, DBLP:journals/corr/abs-2407-15850} but can also be treated thanks to existing annotations such as speaker names from movie transcripts~\citep{DBLP:journals/corr/abs-2005-08646}.
Unlike previous approaches, we perform character identification and video captioning in one go without relying on face annotations from IMDb or training any new algorithms, by prompting a VLM on the audio, video and transcripts (see Section~\ref{subsec:clip_captioning}).

\paragraph{Summary Evaluation}
\label{subsec:summ_eval}
Common strategies for evaluating artificial summaries include word n-gram comparisons such as ROUGE~\citep{lin-2004-rouge} or METEOR~\citep{DBLP:conf/acl/BanerjeeL05}, neural-based evaluation like BertScore~\citep{DBLP:conf/iclr/ZhangKWWA20}, factual consistency evaluation~\citep{DBLP:journals/tacl/LabanSBH22, DBLP:conf/emnlp/KryscinskiMXS20, DBLP:conf/acl/MaynezNBM20, liu-2021-controllable,DBLP:conf/eacl/KrishnaBKIDCL23,liu-2022-entity} or QA-based evaluation~\citep{DBLP:conf/acl/DurmusHD20, DBLP:conf/naacl/FabbriWLX22}.

Most of the above metrics have been widely used in multimodal summary evaluation~\citep{DBLP:conf/eacl/PapalampidiL23, DBLP:conf/acl/MahonL24}.
Yet, they have not been designed to assess the multimodal richness of summaries.
We therefore propose a metric,~\textsc{MFactSum}, specifically designed to evaluate the multimodal content in video and text summaries~(see Section~\ref{sec:MFactSum}).

While there has been some effort in balancing and/or aligning modalities in video summarization models~\citep{DBLP:conf/cvpr/ShenYWYEW22, DBLP:journals/corr/abs-2404-12353, DBLP:conf/emnlp/LiuSYZX20, DBLP:journals/tmm/LinHCLHHL24} and even making their output interpretable with respect to every modality~\citep{DBLP:journals/corr/abs-1812-02872}, there has been, to our knowledge, no specific work on developing balanced summarization metrics for the multimodal setting.

In addition, while some progress has been made in developing metrics for image and text summarization~\citep{DBLP:journals/corr/abs-2402-11414, DBLP:conf/emnlp/WanB22, DBLP:conf/emnlp/HesselHFBC21, DBLP:conf/sdm/Zhang0GL23, DBLP:conf/emnlp/ZhuLL0ZZ18, DBLP:conf/aaai/ZhuZZLZL20}, no such metrics have been proposed yet for videos to the best of our knowledge.

\section{Screenplay Generation and Summarization} \label{sec:summ_model}
We describe our zero-shot multimodal summarization pipeline in Fig.\ref{fig:pipeline}, which takes the whole video and transcript as input. The first step is to select and extract all the clips of interest from the whole video~(Section \ref{subsec:clip_extraction}). We then generate captions for each clip using a VLM while identifying the main characters appearing in them~(Section \ref{subsec:clip_captioning}). We build the screenplay by aligning the resulting clip captions in time together with the transcripts. We finally feed the screenplays to an LLM for summarization~(Section \ref{subsec:screenplay_summarization}).

\subsection{Clip Selection}
\label{subsec:clip_extraction}

Selecting important clips for long video summarization is a challenging task. When it comes to VideoQA, some approaches~\citep{DBLP:conf/nips/Yu0YB23} train a model to predict the most important keyframes given the question to answer.

As we want to keep our pipeline training-free, we select all video clips that occur during a pause in the dialogue when no speech is detected. We ensure that all such video clips are extended to a minimum duration of 10 seconds to give enough context for further captioning.

Our clip selection strategy has two motivations:
\begin{itemize}[leftmargin=*]
    \item Silent scenes from a video often highlight key visual moments and actions impacting the episode storyline. For instance, in \textit{As the World Turns}, a deeply moving moment occurs when Noah inserts the CD of his own movie into a computer~(Figure~\ref{fig:pipeline}).
    \item In Audio Description, the narrator's interventions are usually placed during such breaks in the dialogue suggesting the importance these moments have in the unfolding of an episode.
\end{itemize}

We further validate our proposed criterion in Section~\ref{sec:hhhhuman}. With this criterion, we reduce the cost of our experiments on SummScreen3D (Section~\ref{sec:experiments}) by using about~25\% of the entire video as input.

\subsection{Clip Captioning with Character Names} \label{subsec:clip_captioning}

We feed all video clips to a VLM for captioning. We also provide the corresponding transcripts from SummScreen3D so that the model can reidentify the characters present in the video clips and explicitly state their names in the produced captions. This is possible because each transcript line contains the name of the speaker. The model can therefore easily deduce which characters appear at a given time in the clips by matching each transcript line with the corresponding audio and video. We further validate the performance of our character identification in Section~\ref{sec:hhhhuman}. Detailed captioning prompts are provided in Appendix~\ref{subsec:clip_captioning_prompt}.

\subsection{Screenplay Summarization} \label{subsec:screenplay_summarization}

The screenplay is built by interleaving the clip captions generated in Section~\ref{subsec:clip_captioning} and the transcript utterances together in time. To enhance clarity for the LLM, we prepend \texttt{`Caption:'} to every clip caption within the screenplay. We end up with a document where every clip caption has been inserted at the correct timestamp in the transcripts.
We further summarize those screenplays with a LLM, using a prompt specifically tailored for the multimodal task, asking the model to pay attention to important visual cues like characters' actions or scene locations (see Appendix~\ref{subsec:screenplay_summarization_prompt}).

\section{Multimodal Summary Evaluation with~\textsc{MFactSum}} \label{sec:MFactSum}
When it comes to movie and TV show summarization, the reference summaries are often imbalanced~(see Appendix~\ref{section:example_summaries}) as they usually contain more information regarding what can be heard or read from the transcripts (text modality) than what can be seen (vision modality). Due to this imbalance, most of the evaluation metrics in section~\ref{subsec:summ_eval}, like ROUGE and factual consistency metrics, do not account well for multimodal understanding.

Our proposed metric,~\textsc{MFactSum}, assesses the effectiveness of a multimodal summarization system in balancing and integrating information from both the video and transcripts.~\textsc{MFactSum} therefore gives a new insight on multimodal summary evaluation that cannot provide most traditional summarization metrics. Such a metric is essential to properly assess the inclusion in the summary of all the important video information absent from the transcripts such as performed actions (e.g. \texttt{`Brooke kisses Ridge'}), emotions (e.g. \texttt{`Beth is hysterical'}) or settings (e.g. \texttt{`sitting in a wheelchair'}).

\begin{figure}[hbt]
\begin{center}
\includegraphics[width=0.48\textwidth]{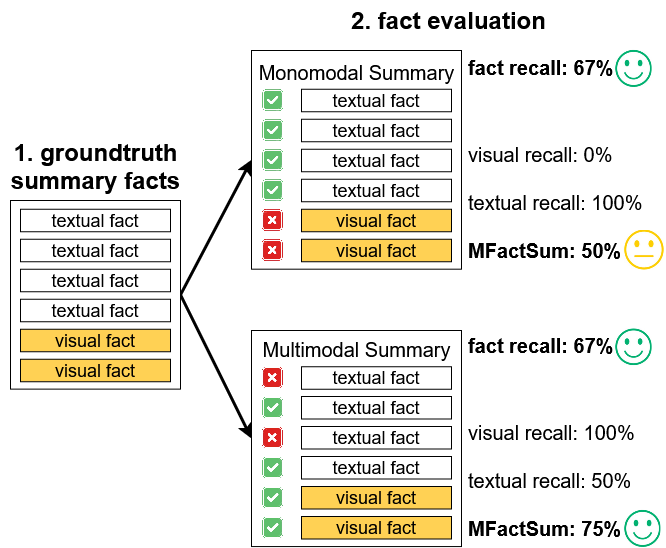}
\end{center}
\caption{\textbf{For the same number of recalled facts,~\textsc{MFactSum} favors the multimodal summary}. 1) We identify all the visual and textual facts within the groundtruth summary. To be consistent with our study dataset, we choose a groundtruth summary with an uneven distribution of textual and visual facts. 2) Although~\textsc{PRISMA}'s fact recall score is the same for both the multimodal and monomodal summaries, our metric~\textsc{MFactSum} favors the multimodal summary over the monomodal one.}
\label{fig:mfactsum_fig}
\end{figure}


\subsection{Metric Design} \label{sec:metric_design}
Our metric is a factual consistency metric similar to Factscore~\citep{DBLP:conf/emnlp/MinKLLYKIZH23} and~\textsc{PRISMA}~\citep{DBLP:conf/acl/MahonL24}. 
\textsc{PRISMA} decomposes the groundtruth summary into its own facts, each fact conveying a single piece of information (roughly equivalent to a simple clause, see examples in Appendix~\ref{subsubsec:face_extraction_prompt}). Completeness of a summary is then evaluated with a "fact recall", i.e. the ratio of facts from the groundtruth summary being supported by the predicted summary.

As opposed to~\textsc{PRISMA}, our metric is not biased towards a single modality as it gives equal importance to visual facts (facts referring to the video) and textual facts (facts referring to the transcripts) in the recall computation. We provide an example comparison of the two metrics in Figure~\ref{fig:mfactsum_fig}.~\textsc{MFactSum} ultimately provides deeper insights into how well an approach understands a video from the perspective of what can be heard (text modality) and what can be seen (vision modality).

Unlike~\textsc{PRISMA}, which also considers fact precision for the relevance of a summary, our metric~\textsc{MFactSum} is recall-based. This is because our study focuses on evaluating the unimodal bias in summarization systems. In other words, we assess whether or not a summary recalls relevant information from both modalities.

\subsection{Metric Computation} \label{subsec:metric_computation}

Our metric computation involves three different steps: \textbf{Fact Identification}, \textbf{Visual Fact Classification} and \textbf{Fact Evaluation}. The only step that is kept unchanged compared to~\textsc{PRISMA} design is the identification of facts.

\textbf{Fact Identification} We split the groundtruth summary into its own sentences. We further identify the facts within each sentence by prompting an LLM in a few-shot setting. Our few-shot examples can be found in Appendix~\ref{subsubsec:face_extraction_prompt}. At this stage, we end up with a list of groundtruth summary facts on which we can later perform our evaluations.

\textbf{Visual Fact Classification} We classify whether a fact from the groundtruth is~\texttt{Visual} (related to the video) or~\texttt{Textual} (related to the dialogue between the characters). We propose two simple steps to separate~\texttt{Visual} from~\texttt{Textual} facts.

\begin{enumerate}
    \item A fact is classified as~\texttt{Visual} if it cannot be inferred from the transcripts alone. We therefore prompt the LLM in zero-shot to answer whether the fact is supported by the transcripts. Our prompt is given in Appendix~\ref{subsubsec:visual_fact_prompt}.
    \item We manually hand-label a list of facts in~\ref{subsubsec:visual_fact_prompt} as~\texttt{Visual} or~\texttt{Textual} and use them as few-shot examples for the classification task.
\end{enumerate}

Each of the above steps is performed independently, and both must classify a fact as~\texttt{Visual} for it to be considered as such. In all other cases, the fact is classified as~\texttt{Textual}. We further validate our classification method in Section~\ref{sec:hhhhuman}.

\textbf{Fact Evaluation} In~\textsc{PRISMA}, the fact recall~(\texttt{fact-rec}) is defined by the ratio of facts from the groundtruth summary that are supported by the predicted summary.

Instead, we compute the recall for both visual and textual facts separately. This allows us to define~\textsc{MFactSum} as the average of visual recall~(\texttt{vis-rec}) and textual recall~(\texttt{text-rec}).
\[
    \textsc{MFactSum} = \frac{\texttt{vis-rec} + \texttt{text-rec}}{2}\,.
\]
Doing the mean at this stage sets the same importance to vision and text modalities in the metric computation.

\section{Experiments}
\label{sec:experiments}

\subsection{Datasets}
\label{subsec:datasets}

SummScreen3D is a video-to-text summarization dataset~\citep{DBLP:conf/eacl/PapalampidiL23} of 5421 episodes of about~30 to~60 minutes each from famous soap operas (As the World Turns, The Bold and the Beautiful,~\ldots). It includes rather long transcripts (about 6K tokens on average), videos and multiple human-annotated summaries for each episode. The validation and test splits contain 296 episodes each.

SummScreen3D is a simple extension of SummScreen~\citep{DBLP:conf/acl/ChenCWG22} into a multimodal summarization dataset by adding full-length videos to the already existing transcripts and reference summaries. The summaries in SummScreen3D are also highly multimodal as they contain information referring to both the episode video and transcripts.

Aligning in time every line from the transcripts to its corresponding frames from the video is crucial for an accurate multimodal understanding of a TV show episode and for generating our screenplays. We rely on previous work~\citep{DBLP:conf/acl/MahonL24} to achieve this.

\subsection{Evaluation Metrics}

We report in Table~\ref{tab:main-results-MFactSum} the multimodal performance based on our new metrics (\texttt{vis-rec},~\texttt{text-rec} and~\textsc{MFactSum}) presented in Section~\ref{sec:MFactSum}. For comparison, we report the simple fact recall score~(\texttt{fact-rec}) as defined in~\textsc{PRISMA} in which no reweighting of visual and textual facts is performed. We use the lighter model Gemini 1.5 Flash as the base model for all our fact-based evaluations.

We also include the average ROUGE-1 (r1), ROUGE-2 (r2) and ROUGE-Lsum (rlsum) as given by the python-rouge package and include two additional metrics, METEOR and BertScore, in Appendix~\ref{app:meteor_bertscore} for further comparison.

As multiple reference summaries are provided for each episode in SummScreen3D, we take the maximum ROUGE score against all references. When doing our fact-based evaluation, we only use the groundtruth summaries from~\texttt{soap\_central} found in SummScreen3D, as they are longer on average and thus more likely to contain visual facts. More precisely, we identify~20 visual facts on average in those summaries making them about~14.5\% of the total number of facts. We filter out the~28 episodes from the test set for which no~\texttt{soap\_central} summary is provided.

We always report the average word count~(\texttt{avg-len}) of a system's summaries in Table~\ref{tab:main-results-MFactSum}, as we are aware that both visual and textual recall can increase with summary length. When comparing any two summarization systems in Section~\ref{subsec:results}, we therefore always make sure that their summary lengths are comparable.

\subsection{Implementation Details}

We generate screenplays for all the~296 episodes from SummScreen3D test split using the pipeline described in Section~\ref{sec:summ_model}. We use either Gemini 1.5 Pro~\citep{DBLP:journals/corr/abs-2403-05530} or Qwen2-VL-72B~\citep{Qwen2VL} as both the captioning and screenplay summarization models. We pick up those two models as they are currently among the best performing VLMs according to multiple long-form video benchmarks~\citep{DBLP:journals/corr/abs-2405-21075, DBLP:conf/nips/MangalamAM23, DBLP:journals/corr/abs-2406-08035} and also for their long context abilities.
We also run various baselines (see Section~\ref{sec:comparison_models}) on the same split for comparison. Due to the high API costs, we only provide results from a single run.

\begin{table*}
\center
\setlength{\tabcolsep}{2pt}
\resizebox{0.86\textwidth}{!}{
 \begin{tabular}{@{}cccc|cccc|c@{}}
\toprule & vis-rec & text-rec & \textsc{MFS} & fact-rec & r1 & r2 & rlsum & avg-len \\
\midrule
\multicolumn{1}{c}{\textbf{multimodal baselines}} & & & & & & & & \\
\multicolumn{1}{l}{Modular-Kosmos~\citep{DBLP:conf/acl/MahonL24}} & 7.39 & 19.56 & 13.48 & 17.90 & 44.86 & \textbf{11.83} &
42.97 & 314.0 \\
\multicolumn{1}{l}{VLog} & 7.77 & 15.66 & 11.72 & 14.62 & 25.99 & 3.11 & 24.69 & 314.0 \\
\midrule
\midrule
\multicolumn{1}{l}{Qwen2-VL-72B (no video)} & 16.16 & 38.0 & 27.08 & 35.0 & 40.6 & 8.97 & 39.28 & 718.5 \\
\multicolumn{1}{l}{Qwen2-VL-72B (video)$^{*}$} & 23.69 & 37.50 & 30.60 & 35.61 & 38.06 & 8.11 & 36.95 & 889.0 \\
\multicolumn{1}{l}{\textit{Screenplay Summary (Qwen2-VL-72B)}} & 24.43 & 35.45 & 29.94 & 33.93 & 36.50 & 7.23 & 35.51 & 749.5 \\
\midrule
\multicolumn{1}{l}{Gemini 1.5 Pro (no video)} & 21.54 & 42.44 & 31.99 & 39.52 & 41.52 & 9.04 & 40.06 & 573.9 \\
\multicolumn{1}{l}{Gemini 1.5 Pro (video)$^{*}$} & 27.48 & 43.00 & 35.24 & 40.87 & \textbf{46.67} & 11.77 & \textbf{44.99} & 688.3 \\
\multicolumn{1}{l}{\textit{Screenplay Summary (Gemini 1.5 Pro)}} & \textbf{33.04} & \textbf{45.12} & \textbf{39.08} & \textbf{43.53} & 40.23 & 8.57 & 38.82 & 601.1 \\
\bottomrule
\end{tabular}
}
\caption{\textbf{Evaluation results on SummScreen3D.} We report the visual recall~(\texttt{vis-rec}), textual recall~(\texttt{text-rec}) and~\textsc{MFactSum} denoted as~\textsc{MFS}. For comparison, we also include ROUGE-1 (r1), ROUGE-2 (r2), ROUGE-Lsum (rlsum) and the simple fact recall~(\texttt{fact-rec}). The average summary word count is denoted by~(\texttt{avg-len}). Best results are in \textbf{bold}. $^*$ indicates the VLM is prompted on the full video and transcripts in an end-to-end fashion using the maximum number of frames allowed by the API.} \label{tab:main-results-MFactSum}
\end{table*}

\subsection{Comparison Baselines} \label{sec:comparison_models}

We compare our approach to existing works from the literature including finetuned models on SummScreen3D and zero-shot pipelines for long video understanding.
As the video-to-text summarization of movies and TV shows is a recent task, very few finetuned end-to-end approaches have been proposed.
In addition, we do not compare to existing works on Audio Description~\citep{DBLP:journals/corr/abs-2407-15850, DBLP:journals/corr/abs-2405-00983, DBLP:conf/cvpr/ZhangLYWLL0W24} because they operate on specific selected video segments rather than the whole video as input.

For that reason, we also include state-of-the-art open and private VLMs such as Gemini 1.5 Pro and Qwen2-VL-72B as comparison baselines in order to perform the summarization task end-to-end. We also tried other VLMs~\citep{DBLP:journals/corr/abs-2412-04468, DBLP:journals/corr/abs-2406-07476, DBLP:journals/corr/abs-2404-16821} but did not include them in our experiments as we found them unsuitable for end-to-end analysis of hour-long videos and transcripts.

\textbf{Modular-Kosmos}~\citep{DBLP:conf/acl/MahonL24} A multimodal approach finetuned on SummScreen3D. It decomposes an episode into its own scenes to further perform summarization and generate video captions for each of them. A higher-level BART~\citep{DBLP:conf/acl/LewisLGGMLSZ20} is then finetuned to fuse both scene summaries and video captions into one global summary for the entire episode.

\textbf{VLog} An open-source tool\footnote{\url{https://github.com/showlab/VLog}} that converts a video into a long document thanks to vision captioners and a speech recognition model. The document is later fed to ChatGPT~\citep{openai2023chatgpt} to start a conversation. To further generate the summaries with VLog, we use the prompt in Appendix~\ref{subsec:vlog_prompt} that is very similar to the one used for summarizing our own screenplays.

\textbf{Gemini 1.5 Pro and Qwen2-VL-72B (video)}
We prompt both models on full videos and transcripts in an end-to-end fashion for multimodal summarization. We extract the maximum number of frames from the videos that are allowed by each model's API which is~1 frame per second for Gemini\footnote{\url{https://aistudio.google.com/}} and~250 frames for Qwen2-VL\footnote{\url{https://www.alibabacloud.com/en/product/modelstudio}}. For fairness, we use a prompt (see Appendix~\ref{subsec:vlm_prompt}) that is very similar to the one we used for our own screenplay summarization approach as we assume it leads to richer multimodal summaries.

\textbf{Gemini 1.5 Pro and Qwen2-VL-72B (no video)}
We generate summaries from the transcripts alone using each model's API. We provide the prompt for this setup in Appendix~\ref{subsec:simple_summ_prompt}.

\begin{table*}
\setlength{\tabcolsep}{2pt}
\center
\resizebox{0.80\textwidth}{!}{
\begin{tabular}{@{}cccc|cccc|c@{}}
\toprule
 & vis-rec & text-rec & \textsc{MFS} & fact-rec & r1 & r2 & rlsum & avg-len\\
\midrule
\multicolumn{1}{l}{w/o handcrafted prompt} & 20.55 & 40.66 & 30.61 & 37.96 & \textbf{40.91} & \textbf{8.78} & \textbf{39.54} & 609.9 \\
\multicolumn{1}{l}{w/o character ident.} & \textbf{34.35} & 43.32 & 38.84 & 42.14 & 39.97 & 8.26 & 38.58 & 576.5 \\
\midrule
\multicolumn{1}{l}{\textit{Screenplay Summary (Gemini 1.5 Pro)}} & 33.04 & \textbf{45.12} & \textbf{39.08} & \textbf{43.53} & 40.22 & 8.65 & 38.90 & 601.1 \\

\bottomrule
\end{tabular}
}
\caption{\textbf{Ablation results for our screenplay summarization pipeline using Gemini 1.5 Pro as the based model.} We report the visual recall~(\texttt{vis-rec}), textual recall~(\texttt{text-rec}) and~\textsc{MFactSum} denoted as~\textsc{MFS}. For comparison, we also include ROUGE-1 (r1), ROUGE-2 (r2), ROUGE-Lsum (rlsum) and the simple fact recall~(\texttt{fact-rec}). The average summary word count is denoted by~(\texttt{avg-len}). Best results are in \textbf{bold}.} \label{tab:abl-results}
\end{table*}

\subsection{Results}
\label{subsec:results}
\textbf{Our screenplay summaries are more multimodal than VLM summaries.}
Our screenplay summaries produced with Gemini 1.5 Pro recall about 20\% more visual facts than Gemini 1.5 Pro prompted end-to-end on the full videos and transcripts, while still maintaining a comparable ability to recall textual information (Table~\ref{tab:main-results-MFactSum}). We observe a similar trend, to a lesser extent, when substituting Gemini 1.5 Pro for Qwen2-VL-72B.

This suggests that, for the task, relying on our screenplays as a multimodal representation, allows us to outperform a VLM model prompted end-to-end.
Note that this improvement in terms of visual recall is not due to a difference in summary length~(\texttt{avg-len}) as our screenplay summaries are actually shorter.

In Appendix~\ref{app:output_summ}, we compare the amount of visual content retrieved by the different tested models in their summary for a single episode.

\textbf{Traditional metrics can overlook the inclusion of both modalities into generated summaries.}
Traditional summarization metrics like ROUGE in Table~\ref{tab:main-results-MFactSum}, even METEOR and BertScore in Appendix~\ref{app:meteor_bertscore}, are often surprisingly higher for summaries that are less multimodal. In particular,
\begin{itemize}
\item Our screenplay summaries are~20\% more visual than Gemini 1.5 Pro prompted end-to-end on the full videos, while they are still about 6 ROUGE points behind.
\item A finetuned model like Modular-Kosmos has better ROUGE scores than most other approaches. Yet, its visual recall is the lowest of all compared to its textual recall, suggesting that its multimodal ability is weak.
\end{itemize}

This phenomenon can be explained by the fact that a multimodal summary is inherently more abstractive than a monomodal one, which ultimately negatively affects all the metrics that rely on exact lexical matching such as ROUGE or METEOR.

\textbf{~\textsc{MFactSum} better reflects multimodality in summaries.}
We compare across all metrics~(Table~\ref{tab:main-results-MFactSum}) the scores between a multimodal summary like ours and a monomodal summary (no video as input). The conclusions here are the same for both Gemini 1.5 Pro and Qwen2-VL.

When no video is used as input, both~\textsc{MFactSum} and the visual recall experience the largest drop. On the other hand, metrics such as the simple fact recall~(\texttt{fact-rec}) either drop a little or even increase like ROUGE.
This means that~\textsc{MFactSum} better assesses the multimodal ability of a summarization system.

Note that the visual recall found for the monomodal summary is however greater than~0 (Table~\ref{tab:main-results-MFactSum}).
We attribute this to potential inaccuracies in fact evaluation, as some facts may be ambiguous in relation to the overall context of an episode.

\subsection{Ablations}
\label{sec:ablations}
We perform various ablations and draw the following conclusions from our results in Table~\ref{tab:abl-results}. 

\textbf{Instruction Tuned LLMs such as Gemini do not naturally output multimodal summaries.}
Simply asking the LLM to summarize the screenplay is not enough to capture its multimodal content. To demonstrate this point we replace our own handcrafted prompt from Section~\ref{subsec:screenplay_summarization} with a much simpler prompt that only asks to summarize (see Appendix~\ref{subsec:simple_summ_prompt}). We discover that, by doing so, the visual recall drops by about~40\%~(Table~\ref{tab:abl-results}).
Note that this drop is not caused by a difference in summary length~(\texttt{avg-len}) between the two compared systems. This suggests that Instruction Tuned models such as Gemini 1.5 Pro are naturally biased into generating monomodal summaries and need to be specifically prompted to retrieve important visual information from the screenplays. Note that this bias can also be found in a finetuned model such as Modular-Kosmos (see Appendix~\ref{section:example_summaries}).

\textbf{Character identification has little impact on our pipeline performance.} Removing the character identification part from our pipeline has little impact on both ROUGE and factual consistency metrics~(Table \ref{tab:abl-results}). One possible reason is that character identities can often be inferred from the screenplay itself, thanks to the dialogue context surrounding each clip caption.
Since our screenplays aim to provide a comprehensive representation of an episode, we decide to still include the character identification component within our main pipeline.
The prompts used for this ablation are in Appendix~\ref{subsec:clip_captioning_prompt}.

\begin{table*}[htbp]
\small
\center
\setlength{\tabcolsep}{2pt}
\resizebox{0.75\textwidth}{!}{
 \begin{tabular}{@{}ccccc|c@{}}
\toprule & Episode 1 & Episode 2 & Episode 3 & Episode 4 & Average/Total \\
\midrule
\texttt{Visual} Fact Precision (\%) & 90.0 & 100 & 86.67 & 93.65 & 92.58 \\
\texttt{Textual} Fact Precision (\%) & 84.92 & 89.47 & 81.19 & 87.60 & 85.80 \\
\texttt{Visual} Fact Recall (\%) & 61.04 & 50.0 & 40.63 & 50.0 & 50.42 \\
\texttt{Textual} Fact Recall (\%) & 97.13 & 100 & 97.62 & 99.07 & 98.46 \\
\midrule
\texttt{Visual} Fact Count & 77 & 12 & 32 & 30 & 151 \\
Fact Count & 257 & 63 & 116 & 137 & 573 \\
\bottomrule
\end{tabular}
}
\caption{\textbf{Human evaluation of the visual fact classification.} We provide the estimated Precision and Recall for both~\texttt{Visual} and~\texttt{Textual} facts for each episode and aggregated over all episodes. We also report the total number of facts per episode as well as~\texttt{Visual} facts found by the human evaluator.} \label{tab:human_vis_class}
\end{table*}

\subsection{Human Evaluation} \label{sec:hhhhuman}

We conduct a human evaluation of various components of our summarization pipeline and metric. We provide here the details for the evaluation of the~\textsc{MFactSum} metric and in Appendix~\ref{sec:human_val} for the other components of our summarization pipeline. All evaluations are performed by one of the co-authors on the same 4 episodes, sampled from the SummScreen3D test set. Our human evaluation is statistically significant as each episode comprises numerous instances of what is evaluated (573 facts, 174 video clips in total).

\begin{itemize}
    \item Episode 1 is the episode from \textit{As the World Turns} aired on 01-05-2010.
    \item Episode 2 is the episode from \textit{Guiding Light} aired on 01-25-2005.
    \item Episode 3 is the episode from \textit{the Bold and the Beautiful} aired on 05-29-2006.
    \item Episode 4 is the episode from \textit{the Bold and the Beautiful} aired on 06-12-2006.
\end{itemize}

\textbf{MFactSum} Our metric robustly identifies visual facts from groundtruth summaries, achieving an accuracy of about~86.3\% on~573 facts. Note that we only evaluate the classification of visual facts as it is the main novelty of~\textsc{MFactSum}.

As described in Section~\ref{subsec:metric_computation}, we extract all the groundtruth summary facts and automatically classify them as either~\texttt{Visual} (video related) or~\texttt{Textual} (transcript related) using our metric. We also ask our human evaluator to independently label each fact as~\texttt{Visual} or~\texttt{Textual}.

We then study how the classification performed by our metric compares against those human annotations. Over the~4 episodes (573 facts) on which we perform human evaluation, our metric performs visual fact classification with an accuracy of~86.3\%.

We further study the precision and recall for each class in Table~\ref{tab:human_vis_class}. We are able to achieve a high average precision on both~\texttt{Visual}~(92.58\%) and~\texttt{Textual}~(85.80\%) facts. This means our evaluation metric is robust as we are able to clearly distinguish between~\texttt{Visual} and~\texttt{Textual} facts.

While the precision for~\texttt{Visual} facts is very high~(92.58\%), the corresponding recall is lower~(50.42\%). This shows the task difficulty. In the meantime, leveraging the whole video in order to improve the classification performance is currently out of reach due to VLMs' limitations in long video understanding.
By keeping a high precision for~\texttt{Visual} facts, at the expense of a lower recall, one can ensure the robustness of~\textsc{MFactSum} by not introducing noise in later stages of the metric computation.

\textbf{Clip Selection} Our selected video clips (Section~\ref{subsec:clip_extraction}) capture approximately two thirds of the visual information present in groundtruth summaries, which is about 1.9 times more than what is achieved using randomly selected clips (Appendix~\ref{app:human_val_clip_selection}).

\textbf{Character Identification} We found a~77.4\% overlap with the human annotator in identifying characters across 174 video clips (Appendix~\ref{app:human_val_char_ident}). This result is on par with prior work.

\section{Conclusion}
In this paper, we introduce a zero-shot video-to-text summarization approach that constructs a multimodal screenplay of an episode, effectively integrating important visual cues, transcripts, and character information.
Unlike previous approaches, we recognize characters in the video while producing video captions simultaneously using only the audio, video, and transcripts as input, avoiding the need for extra face annotations (IMDb).
We also propose a new multimodal metric,~\textsc{MFactSum}, which better reflects the multimodal fidelity of summaries than traditional metrics according to our experimental results. Evaluation with~\textsc{MFactSum} shows that LLMs and finetuned models are naturally biased and tend to include less visual content in their summaries. In contrast, using our screenplay representation leads to summaries that not only incorporate 20\% more relevant visual information than state-of-the-art vision-language models like Gemini 1.5 Pro but also require 75\% less of the video as input.

Future work could include expanding our evaluation metrics to other multimodal generative tasks and domains, as well as exploring methods to reduce modality biases through better representation learning and alignment.

\section*{Limitations}
The motivation behind our proposed clip selection strategy comes from the audio description of movies or TV shows~\ref{subsec:clip_extraction}. Different strategies should be explored to identify key moments for other video domains while keeping the overall inference cost low.

Despite the effectiveness of our evaluation approach, we identify two main challenges that still need to be addressed.
\begin{itemize}
\item \textbf{Fact Evaluation:} LLMs might still generate some errors in the evaluation as the facts can sometimes be ambiguous with respect to the whole context of the episode. 
\item \textbf{Visual Fact Classification:} Leveraging the episode video for identifying visual facts is currently out of reach as current VLMs still produce inaccuracies when prompted on long videos (e.g. long video question answering).
\end{itemize}

Finally, the financial cost per episode of generating our screenplay summaries and evaluating them using~\textsc{MFactSum} is approximately~$1.25$~\$ per episode. We also estimate the total cost of producing all the results presented in this paper to be approximately~1000~\$. In particular, our fact-based evaluation with~\textsc{MFactSum} involves multiple LLM queries with repeated long contexts, further increasing inference costs.

\section*{Acknowledgments}
We thank the anonymous reviewers for their feedback. This research is supported by the National Research Foundation, Prime Minister’s Office, Singapore under its Campus for Research Excellence and Technological Enterprise (CREATE) programme. Any opinions, findings and conclusions or recommendations expressed in this material are those of the authors and do not reflect the views of the National Research Foundation, Singapore.

\bibliography{anthology,custom}
\bibliographystyle{acl_natbib}

\appendix
\newpage

\section{Prompts}

\subsection{Clip captioning Prompt} \label{subsec:clip_captioning_prompt}

Below are the prompt templates used for generating clip caption. The video clips are processed by the VLM at one frame per second (1 fps).

Subsequently, we always ask the model to summarize its produced output into a few sentences.

\subsubsection{Clip Captioning Prompt with Character Identification}

For the prompt with character identification, we additionally provide the transcripts as input to the model. In that case, the model can internally match every line from the transcripts to the audio of the video in order to infer the names of the characters appearing in every frame.

\begin{tcolorbox}[colback=white, colframe=green!20, left=2pt,  coltitle=black,  title=\textbf{}]

{\color{blue} <VIDEO CLIP>}\\

Here are the transcripts for the corresponding video:

{\color{blue} <CLIP TRANSCRIPTS>}\\

Describe what is happening in the video in all the details.

Explicitly state the names of the characters in your description when possible.

\end{tcolorbox}

\subsubsection{Clip Captioning Prompt without Character Identification} \label{app:clip_caption_no_char}

The prompt without character identification is used essentially for the ablation study (see Section~\ref{sec:ablations}).
Also, since Qwen2-VL does not have access to the audio, we do not perform character identification with this model and always use the prompt below.

\begin{tcolorbox}[colback=white, colframe=green!20, left=2pt,  coltitle=black,  title=\textbf{}]

{\color{blue} <VIDEO CLIP>}\\

Describe what is happening in the video in all the details.

\end{tcolorbox}

\subsection{Summarization Prompts}

We provide here the prompts we used for generating summaries in our experiments. We write in {\color{red}red} the part of the prompt that varies across the tested models.

\subsubsection{Screenplay Summarization Pipeline} \label{subsec:screenplay_summarization_prompt}

\begin{tcolorbox}[colback=white, colframe=green!20, left=2pt,  coltitle=black,  title=\textbf{}]
Summarize every single existing subplot from the above dialogue. For each subplot, include throughout your summary any important visual detail or information about character actions, interactions, scene location {\color{red}that you may find in the Video Captions.}
\end{tcolorbox}

\subsubsection{VLog} \label{subsec:vlog_prompt}

\begin{tcolorbox}[colback=white, colframe=green!20, left=2pt,  coltitle=black,  title=\textbf{}]
Summarize every single existing subplot from the above dialogue. For each subplot, include throughout your summary any important visual detail or information about character actions, interactions, scene location.
\end{tcolorbox}

\subsubsection{Gemini 1.5 Pro and Qwen2-VL (video)} \label{subsec:vlm_prompt}

\begin{tcolorbox}[colback=white, colframe=green!20, left=2pt,  coltitle=black,  title=\textbf{}]
Summarize every single existing subplot from the above dialogue. For each subplot, include throughout your summary any important visual detail or information about character actions, interactions, scene location {\color{red} that you may pick up from the video frames and provided images.}
\end{tcolorbox}

\subsubsection{Simple Summarization Prompt}
\label{subsec:simple_summ_prompt}

\begin{tcolorbox}[colback=white, colframe=green!20, left=2pt,  coltitle=black,  title=\textbf{}]
Summarize every single existing subplot from the above dialogue.{\color{red} Your summary should be very complete.}
\end{tcolorbox}

\subsection{Multimodal Summary Evaluation Prompts}

\subsubsection{Fact Identification} \label{subsubsec:face_extraction_prompt}

We provide below the few-shot examples and prompts we used for extracting the facts from one groundtruth summary sentence.

\begin{tcolorbox}[breakable, colback=white, colframe=green!20, left=2pt,  coltitle=black,  title=\textbf{}]
Please break down the following sentence into independent facts:

Katie went to Al's diner and reacted to a `Closed' sign on the door.

- Katie went to Al's diner.

- Katie reacted to a `Closed' sign on the door.\\

Please break down the following sentence into independent facts:

Simon ushered Lily in, and she spied a romantic candlelit table just for two.

- Simon ushered Lily in

- Lily spied a romantic candlelit table just for two.\\

Please break down the following sentence into independent facts:

Luke shouted at his lover that the awards were all for him, but Noah shoved the award into Luke's hands and went inside.

- Luke shouted at his lover that the awards were all for him.

- Noah shoved the award into Luke's hands and went inside.\\

Please break down the following sentence into independent facts:

Bridget suggests that perhaps Eric can help them

- Bridget suggests that perhaps Eric can help them\\

Please break down the following sentence into independent facts:

At work at the diner, Simon called Metro to make a dinner reservation, as a customer requests him for service and gives him a hard time making fun of him.

- Simon is at work at the diner.

- Simon called Metro to make a dinner reservation.

- A customer requests Simon for service.

- A customer gives Simon a hard time.

- A customer makes fun of Simon.\\

Please break down the following sentence into independent facts:

Noah and Jack became furious and accused Luke of taking over their work in order to control them.

- Noah and Jack became furious at Luke.

- Noah and Jack accused Luke of taking over their work in order to control them.\\

Please break down the following sentence into independent facts:

At Marone, Taylor pays Nick a visit.

- Taylor and Nick are at Marone.

- Taylor pays Nick a visit.\\

Please break down the following sentence into independent facts:

{\color{blue} <INPUT FACT>}
\end{tcolorbox}

\subsubsection{Visual Fact Classification} \label{subsubsec:visual_fact_prompt}

By definition, a visual fact is a fact that cannot be inferred from the transcripts alone. We therefore separate visual from textual facts by asking the LLM the following question.

\begin{tcolorbox}[colback=white, colframe=green!20, left=2pt,  coltitle=black,  title=\textbf{Fact Evaluation against Transcripts}]
{\color{blue} <TRANSCRIPTS>}\\

Here is a Fact: {\color{blue} <INPUT FACT>}.

Suppose you are given only the above Transcripts.
You do not have access to the Fact.
You want to explain to someone everything that happened in the full transcripts above.
Do you think that your explanation will contain the given Fact?\\

Answer by True or False.
Justify your answer.
\end{tcolorbox}

In addition, we manually hand-label a set of facts as either~\texttt{Visual} or~\texttt{Textual} (as below) and prompt the LLM to classify each given fact in a few-shot setting using the examples below.

\begin{tcolorbox}[breakable, colback=white, colframe=green!20, left=2pt,  coltitle=black,  title=\textbf{Few-shot Visual Fact Classification}]
You are given a list of Facts extracted from a movie.

In what follows, your mission is to tell whether the fact is related to what is seen on the screen or to the conversation between the different characters of the story.

Can the fact be deduced from the conversation between the characters?\\

Fact: There was a `Closed' sign on the door.

Answer: False\\

Fact: Simon ushered Lily in.

Answer: False\\

Fact: Luke shouted at his lover that the awards were all for him.

Answer: True\\

Fact: Noah shoved the award into Luke's hands and went inside.

Answer: False\\

Fact: Molly felt that something was not quite right about the situation.

Answer: True\\

Fact: Simon is at work at Al's diner.

Answer: True\\

Fact: The client's name is Laura and she is frustrated with the service.

Answer: True\\

Fact: Holden notices a look in Molly's eyes that indicates that Molly does not believe Meg.

Answer: True\\

Fact: The TV show is aimed at mothers with children.

Answer: True\\

Fact: Simon is having a hard time dealing with the customer.

Answer: True\\

Fact: There was a noticeable change in Meg's condition.

Answer: True\\

Fact: Tim was making fun of John.

Answer: True\\

Fact: Paul requests the teacher for service.

Answer: True\\

Fact: Lucinda asked a couple of direct questions about Lily being pregnant.

Answer: True\\

Fact: Noah accused Luke of taking over his work in order to control him.

Answer: True\\

Fact: {\color{blue} <INPUT FACT>}

Answer:

\end{tcolorbox}

\subsubsection{Fact evaluation}

We provide below the prompt for evaluating the recall for any generated summary. We use the following question for testing whether a fact is supported by the groundtruth summary.

\begin{tcolorbox}[colback=white, colframe=green!20, left=2pt,  coltitle=black,  title=\textbf{}]
{\color{blue}<TRANSCRIPTS>}\\

Is the Input supported by the above summary?

Input: {\color{blue} <INPUT FACT>}.\\

Answer by True or False. Justify your answer.
\end{tcolorbox}

\begin{table*}[htbp]
\small
\center
\setlength{\tabcolsep}{2pt}
\resizebox{0.65\textwidth}{!}{
 \begin{tabular}{@{}ccccc|c@{}}
\toprule \textbf{\% of visual facts in clips} & Episode 1 & Episode 2 & Episode 3 & Episode 4 & Average/Total \\
\midrule
our clip selection & 77.6 & 66.7 & 53.3 & 73.3 & 67.7 \\
random baseline & 49.0 & 16.7 & 40.0 & 40.0 & 36.4 \\
\midrule
Visual Fact Count & 77 & 12 & 32 & 30 & 151 \\
\bottomrule
\end{tabular}
}
\caption{\textbf{Human evaluation of our clip selection strategy against the random baseline.} We provide the ratio of visual facts retrieved by either clip selection method. We also report the total number of~\texttt{Visual} facts in each episode groundtruth summary.} \label{tab:human_clip_selection}
\end{table*}
\begin{table*}[htbp]
\small
\center
\setlength{\tabcolsep}{2pt}
\resizebox{0.65\textwidth}{!}{
 \begin{tabular}{@{}ccccc|c@{}}
\toprule & Episode 1 & Episode 2 & Episode 3 & Episode 4 & Average/Total \\
\midrule
IoU score (\%) & 81.89 & 74.06 & 69.71 & 83.89 & 77.39 \\
\midrule
Clips Count & 69 & 50 & 25 & 30 & 174 \\
\bottomrule
\end{tabular}
}
\caption{\textbf{Human evaluation of the character identification strategy.} We provide the average IoU score for each episode and aggregated over all episodes. We also report the total number of clips for each episode on which human evaluation is performed.} \label{tab:human_char_ident}
\end{table*}

\section{Human Validations} \label{sec:human_val}

\subsection{Human Validation of the Clip Selection Strategy} \label{app:human_val_clip_selection}
We provide here a human evaluation for the clip selection strategy proposed in Section~\ref{subsec:clip_extraction}.

We study how our clip selection strategy compares to the random baseline. For the random baseline, we randomly select clips from within the whole episode video. For fairness, we also make sure that we select the same number of clips as in our own clip selection strategy and that the total sampled duration is the same.

For each visual fact from the groundtruth summary, we ask our human evaluator to manually check whether the fact is supported by one of the video clips retrieved by either the random baseline or our clip selection strategy.

The results of the human evaluation are given in Table~\ref{tab:human_clip_selection}. Over the 4 episodes (151 visual facts) used for the human validation, our clip selection strategy is found to retrieve about~1.9 times more visual information than the random baseline.

\subsection{Human Validation of the Character Identification} \label{app:human_val_char_ident}
We include here the results of our human evaluation for the character identification strategy proposed in Section~\ref{subsec:clip_captioning}.

As described in Sections~\ref{subsec:clip_extraction} and~\ref{subsec:clip_captioning}, we begin by applying our clip selection algorithm to each episode. For every selected clip, we generate a caption that includes predicted character names. We then ask our human evaluator to independently name all the characters appearing in each video clip and compare to those found in our generated captions.

Following~\citep{DBLP:conf/cvpr/HanBNVXZ24}, we report the IoU score to assess the performance of our method to correctly identify characters in the generated clip captions. The IoU score is given by:
\[
\text{IoU} = \frac{|\mathbf{E}_{\text{pred}} \cap \mathbf{E}_{\text{human}}|}{|\mathbf{E}_{\text{pred}} \cup \mathbf{E}_{\text{human}}|}
\]

where $|\mathbf{E}_{\text{pred}} \cap \mathbf{E}_{\text{human}}|$ is the number of distinct characters correctly identified in the generated clip caption and $|\mathbf{E}_{\text{pred}} \cup \mathbf{E}_{\text{human}}|$ is the total number of distinct characters found by either the human evaluator or in the generated clip caption.

Over the 4 episodes (174 video clips), we show in Table~\ref{tab:human_char_ident} that our generated clip captions share 77.4\% of characters in common with the original video according to our human evaluator. This makes our character identification prompting a reasonable choice for the studied dataset. We found those results to be on par with other strategies used for movies or TV shows as proposed in~\citep{DBLP:conf/cvpr/HanBNVXZ24} and~\citep{DBLP:journals/corr/abs-2407-15850} in which they respectively found a score of 70.8\% and 75.8\% based on 4 movies from the MovieNet dataset~\citep{DBLP:conf/eccv/HuangXRWL20}.

\section{Summary Imbalance and Textual Bias in Finetuned Models}
\label{section:example_summaries}

Table~\ref{tab:gold-bb-summ} shows the gold summary for one episode from the TV show \textit{The Bold and the Beautiful} aired on May 5, 2006. After watching the whole episode, we manually highlight in \hlgr{green} all the visual content present in the groundtruth summary. We notice that visual information, while essential to understanding the episode, is present in only a minority of the groundtruth summary.

The summary produced by Modular-Kosmos (see Table~\ref{tab:modular-bb-summ}) completely ignores all the visual content from the episode only focusing on relating the conversation between the characters.
As shown in Table~\ref{tab:modular-bb-summ}, most of the summary sentences generated by the model are indeed limited to~\texttt{"[Someone] \boxred{says}"},~\texttt{"[Someone]~\boxred{tells}"} or~\texttt{"[Someone]~\boxred{asks}"}.
This may be due to the limited presence of visual information in training summaries, leading finetuned models to focus primarily on dialogue understanding of an episode.

\begin{table*}[htbp]
\setlength{\tabcolsep}{2pt}
\center
\resizebox{0.75\textwidth}{!}{
 \begin{tabular}{@{}ccccc|c@{}}
\toprule
 & METEOR & bert-prec & bert-rec & bert-f1 & avg-len\\
\midrule
\multicolumn{1}{c}{\textbf{multimodal baselines}} \\
\multicolumn{1}{l}{Modular-Kosmos~\citep{DBLP:conf/acl/MahonL24}} & 31.45 & \textbf{83.52} &
\textbf{85.03} & \textbf{84.25} & 314.0 \\
\multicolumn{1}{l}{VLog} & 21.09 & 80.35 & 82.31 & 81.30 & 314.0 \\
\midrule
\midrule
\multicolumn{1}{l}{Qwen2-VL-72B (no video)} & 29.46 & 81.11 & 84.59 & 82.80 & 718.5 \\
\multicolumn{1}{l}{Qwen2-VL-72B (video)$^{*}$} & 21.53 & 80.04 & 82.51 & 81.25 & 889.0 \\
\multicolumn{1}{l}{\textit{Screenplay Summary (Qwen2-VL-72B)}} & 20.79 & 79.48 & 82.42 & 80.92 & 749.5 \\
\midrule
\multicolumn{1}{l}{Gemini 1.5 Pro (no video)} & 31.83 & 82.52 & 84.81 & 83.6 & 573.94 \\
\multicolumn{1}{l}{Gemini 1.5 Pro (video)$^{*}$} & \textbf{33.38} & 81.75 & 84.89 & 83.28 & 688.3 \\
\multicolumn{1}{l}{\textit{Screenplay Summary (Gemini 1.5 Pro)}} & 27.97 & 81.68 & 83.99 & 82.81 & 601.1 \\
\bottomrule
\end{tabular}
}
\caption{\textbf{METEOR and BertScore Evaluation Results on SummScreen3D.} We respectively denote BertScore precision, recall and F1 score by \texttt{bert-prec},~\texttt{bert-rec} and~\texttt{bert-f1}. The average summary word count is denoted by~\texttt{avg-len}. Best results are in \textbf{bold}. $^*$ indicates the VLM is prompted on the full video and transcripts in an end-to-end fashion using the maximum number of frames allowed by the API.} \label{tab:main-results_meteor_bertscore}
\end{table*}

\begin{table*}[htbp]
\setlength{\tabcolsep}{2pt}
\center
\resizebox{0.75\textwidth}{!}{
 \begin{tabular}{@{}ccccc|c@{}}
\toprule
 & METEOR & bert-prec & bert-rec & bert-f1 & avg-len\\
\midrule
\multicolumn{1}{l}{w/o handcrafted prompt} & \textbf{27.97} & \textbf{81.99} & 83.85 & \textbf{82.90} & 609.9 \\
\multicolumn{1}{l}{w/o character ident.} & 27.76 & 81.53 & 83.88 & 82.68 & 576.5 \\
\midrule
\multicolumn{1}{l}{\textit{Screenplay Summary (Gemini 1.5 Pro)}} & \textbf{27.97} & 81.68 & \textbf{83.99} & 82.81 & 601.1 \\
\bottomrule
\end{tabular}
}
\caption{\textbf{Ablation results on METEOR and BertScore for our screenplay summarization pipeline.} We respectively denote BertScore precision, recall and F1 score by \texttt{bert-prec},~\texttt{bert-rec} and~\texttt{bert-f1}. The average summary word count is denoted by~\texttt{avg-len}. Best results are in \textbf{bold}.} \label{tab:ablations_meteor_bertscore}
\end{table*}

\begin{table*}[hbt]
\center
\small
\begin{tabular}{p{15cm}}
\begin{large}
   The gold Summary of \textit{The Bold and the Beautiful} episode (aired 05-05-06)
\end{large} \\
\toprule
\texttt{Ridge continues to beg Brooke to reconsider her decision to leave Forrester as Stephanie continues to voice her opinion. At Marone, \hlgr{Taylor pays Nick a visit}. Nick is \hlgr{still angry} about what Taylor implied when she disclosed that Brooke and Ridge slept together. Taylor tries to apologize and asks if things are all right between Nick and Brooke. Nick tells her that everything is fine and Brooke is quitting her job at Forrester. Taylor is unconvinced that Brooke will be able to let go of either Forrester or Ridge ! Brooke tells Ridge that she cannot fight with Stephanie any longer and that her future is with Nick. \hlgr{After kissing Ridge} and saying that a part of her will always love him, \hlgr{she takes her things and leaves}. \hlgr{Bridget and Dante are at home} discussing Stephanie's interference in the custody of Dino. Bridget suggests that perhaps Eric can help them. \hlgr{Dante worries} about what losing his job would do to his work visa. Bridget convinces him that because they all love Dino, they should be able to work something out. \hlgr{After some wine}, Bridget reveals that she is ready to make love with Dante. \hlgr{As the two were in bed, Dante stops and reaches in the bedside drawer and presents Bridget with an engagement ring} and pops the question. \hlgr{Brooke goes to see Nick at his office} and tells him that she has left Forrester. \hlgr{Nick is pleased}, although Brooke confesses that she hurt Ridge badly by walking out. Nick whisks her off to the Marone jet for a surprise getaway! At Forrester, Ridge \hlgr{angrily} accuses Stephanie of causing all his problems with Brooke. \hlgr{Stephanie is stunned} as Ridge bashes her with a vengeance and then \hlgr{clutching his chest, collapses to the floor!}
}\\ \bottomrule
\end{tabular}
\caption{Gold summary from the SummScreen3D dataset for \textit{The Bold and the Beautiful} episode (aired 05-05-06). Several summaries exist for this episode from different fansites. This summary is from TVMegaSite.} \label{tab:gold-bb-summ}
\end{table*}

\begin{table*}[hbt]
\center
\small
\begin{tabular}{p{15cm}}
\begin{large}
    Modular-Kosmos Summary of \textit{The Bold and the Beautiful} episode (aired 05-05-06)
\end{large} \\
\toprule
\texttt{Bridget and Dante plan to get married in Italy, but Bridget wants to spend more time with Dante. Stephanie \boxred{tells} Bridget that she wants to fire Dante and send him to Italy. Ridge \boxred{tells} Stephanie that he wants Brooke out of Forrester Creations. Brooke \boxred{tells} Nick that she is through fighting, she is moving to Paris. Nick \boxred{tells} her that she has to move out of the office. Nick \boxred{says} that he will not force her to work in Paris, but she \boxred{says} that she needs to move in. Ridge \boxred{says} that Nick is her future, and she has no choice but to leave. Stephanie \boxred{says} that Ridge needs her, but he does not want to accept her. He \boxred{says} that they are better off without her. She \boxred{says} that is not what she wants. She \boxred{tells} him that she does not know how he feels. She does not understand how he could do this to Brooke. He \boxred{tells} her to move on with her life. Taylor apologizes to Nick for suggesting that Ridge and Brooke slept together. Nick \boxred{tells} Taylor that he doesn’t let it affect his relationship with Brooke. Brooke \boxred{tells} Nick that she needs something to cheer her up, and Nick \boxred{tells} her that they are on their way to a tropical island. Ridge tries to convince Brooke to stay at Forrester Creations, but she \boxred{tells} him that she is leaving anyway. Taylor \boxred{tells} Ridge that she does not want to believe that he and Brooke have feelings for each other, but Ridge \boxred{tells} Taylor not to give up on her dreams. Nick and Brooke plan to take off on their trip, but Nick \boxred{tells} Brooke that he has a surprise for her, and that they will be going to a beautiful beach and a fruity drink. Brooke is very happy about Nick’s trip plans, but when she \boxred{asks} where they are going, Nick reveals that he wants to take her to a romantic beach.
}\\ \bottomrule
\end{tabular}
\caption{The summary generated by Modular-Kosmos model for \textit{The Bold and the Beautiful} episode (aired 05-05-06).} \label{tab:modular-bb-summ}
\end{table*}

\section{Additional Metrics} \label{app:meteor_bertscore}
We report in Tables~\ref{tab:main-results_meteor_bertscore} and~\ref{tab:ablations_meteor_bertscore} additional results on METEOR and BertScore metrics. Consistently with our experiments on ROUGE~(Section~\ref{subsec:results}), we always take the maximum score against all references. We respectively leverage the~\texttt{meteor\_score} function from~\texttt{nltk.translate} and the~\texttt{bert\_score} python package.

Similar to what we observed with ROUGE (see Section~\ref{subsec:results}), we found that both metrics are not able to tell the difference between a multimodal and a monomodal summary. Indeed, both METEOR and BertScore in Table~\ref{tab:main-results_meteor_bertscore} were found to be even higher when videos were removed from our screenplay summarization pipeline, regardless of the base model used (Gemini 1.5 Pro or Qwen2-VL-72B).

\section{Models comparison based on visual recall ability} \label{app:output_summ}
We consider the episode from the TV show Guiding Light aired on January 25, 2005, for which we provide the groundtruth summary in Table~\ref{tab:gt-summ}. After watching the whole episode, we compare the summaries produced by our baselines (Section~\ref{sec:comparison_models}) to our own screenplay summary produced with Gemini 1.5 Pro as the base model (see Table~\ref{tab:screenplay_summ_app}).

In all the summaries, we manually highlight in \hlgr{green} the visual information that is supported by the groundtruth, in \hly{yellow}, the visual information that is present in the episode but not mentioned in the groundtruth and in \hlred{red} all the visual information that is hallucinated by the model.

For the chosen example, we notice our screenplay summaries are able to recall (in green) more visual information than any other tested baselines including Gemini 1.5 Pro prompted end-to-end on the full videos and transcripts (see Table~\ref{tab:gemini_end_to_end}).

We also report the visual recall that we automatically compute using Gemini 1.5 Flash for all models' summaries of the chosen episode. The results confirm our qualitative analysis: our screenplay summary successfully retrieves more visual facts than any other tested baseline.

\begin{table*}[hbt]
\center
\small
\begin{tabular}{p{15cm}}

\begin{large}
    Groundtruth Summary of \textit{Guiding Light} episode (aired 01-25-05)
\end{large} \\
\toprule
\texttt{Josh and Reva have gone out to dinner to try to get away from the "situation" at home. But Reva can't help herself, and tries to call Jonathan to check in on him. She doesn't trust the nurse they hired to take care of him while they were out. Her maternal instincts were right! Cassie talked the nurse into trading places with her so she could have Jonathan to herself. She confronts Jonathan about flowers which were sent to her wedding suite, but Jonathan denies everything. Dopey from pain killers and beer, he falls asleep. Cassie takes a close look at his pain killers and just as she pours a few in her hand, Josh and Reva walk in and see her. Reva demands to know what she has done to her son, but Cassie says she has done nothing. Reva is very angry and Josh and Cassie are angry with her. Josh tries to tell her that he is sick of the anguish Jonathan is causing the family. Cassie says her sister is "gone." When Cassie leaves to go home, Josh follows her. He thinks this is a family issue and they will work it out. But Cassie reminds him the problem is Reva thinking of Jonathan as family, too. Indoors, Reva tells Jonathan things will work out, as they always do.}

\texttt{Through some detective work, Harley has found Beth at a spa. She tries to talk with her, but Beth is hysterical. Harley assumes it is because Beth wants to confess that she killed Phillip. But instead, Beth admits the night Phillip was killed, she was going to beg to go with him! Harley is furious when Beth goes on to tell her that she wanted to go with all the children, including Harley's, and that no one would ever have seen the children again. Beth is shocked at herself that she could have considered such a plan, but Harley is crestfallen when she realizes this means she may have killed Phillip. She is inconsolable.}

\texttt{But\ldots back in Springfield, Gus is chatting with Lizzie. She tells him her mother was with her every minute of the evening Phillip was killed. This directly contradicts what Beth has just told Harley. Does Lizzie believe her mother killed her father? Did Lizzie kill her father?}

\texttt{Unfortunately, Alan is very angry at Gus and has followed him to Elizabeth \& Company to have words. Alan overhears a cell phone conversation between Gus and Harley, including when Gus tells Harley the judge would be very angry if he found out she had crossed state lines.}
\\ \bottomrule
\end{tabular}
\caption{Groundtruth summary for the \textit{Guiding Light} episode (aired 01-25-05).} \label{tab:gt-summ}
\end{table*}








\begin{table*}[hbt]
\center
\small
\begin{tabular}{p{15cm}}

\begin{large}
    Screenplay Summary of \textit{Guiding Light} episode (aired 01-25-05)
\end{large} \\
\toprule
\texttt{\textbf{1. Reva and Josh's Relationship Struggles Over Jonathan:}  This subplot centers around the tension between Reva and Josh caused by Jonathan's manipulative behavior. Reva consistently defends Jonathan, even against accusations of drugging Cassie the night before her wedding and other problematic actions.  Josh is increasingly frustrated with Reva's blind loyalty to her son.  They have a tense conversation \hly{in their living room}, where Reva accuses Josh of taking Cassie's side. The argument escalates, with Josh expressing concern about Reva's enabling behavior and Reva feeling isolated and misunderstood.  \hlred{Visual cues like Reva offering a handshake before exiting the room, pausing mid-departure}, and \hlgr{Josh pursuing Cassie onto a dimly lit porch highlight the growing distance between them}.}

\texttt{\textbf{2. Cassie's Quest for Answers and Revenge against Jonathan:} Cassie is deeply disturbed by Jonathan's actions, particularly the implied drugging incident the night before her wedding.  She confronts him multiple times, demanding the truth. Jonathan, \hly{sitting in a wheelchair}, is evasive and mocking.  Cassie's desperation is evident as she initially considers harming him, \hlgr{holding a bottle of pills}.  She later vents her frustration to Reva and Josh, revealing the depth of her anger and hurt.  \hlgr{The visual of Cassie shaking pills into her hand while Jonathan dozes emphasizes her contemplation of revenge}.}

\texttt{\textbf{3. Gus's Determination to Clear Harley's Name:} Gus firmly believes in Harley's innocence in Phillip's murder and is tirelessly pursuing other leads. He clashes with Alan, who believes Harley is guilty and sees Gus's efforts as a betrayal of the family. Gus questions Lizzie about Beth's whereabouts the night of the murder, hoping to uncover information that could exonerate Harley.  The tension between Gus and Alan \hly{plays out in an office setting, near a couch, bookshelf, and lamp}. Gus's desperation is highlighted by \hlred{his abduction of Lizzie, covering her mouth and forcing her towards an exit}.}

\texttt{\textbf{4. Beth and Harley's Shared Secret:}  Beth confesses to Harley her plan to leave the country with Phillip and their children, a plan she abandoned upon arriving at Company.  Harley, in turn, reveals her own turmoil and uncertainty about her actions that night, suggesting she may have been the one who shot Phillip.  \hlgr{The two women share a moment of vulnerability and mutual understanding in a spa-like room}\hly{ with candles and a potted plant}, promising to keep each other's secrets. \hly{Visuals like Beth sitting on the floor sobbing, Harley kneeling beside her in a dark red jacket, and Beth's reflection in the mirror underscore their }\hlgr{emotional distress}.}

\texttt{\textbf{5. The Mystery of Phillip's Murder:}  The overarching plot revolves around the unsolved murder of Phillip. While Harley is the prime suspect, Gus's investigation suggests other possibilities.  Lizzie provides an alibi for Beth, claiming she was with her mother the entire night.  The dialogue heavily focuses on the events of that night, with various characters recalling their actions and whereabouts.  The mystery is fueled by conflicting accounts, withheld information, and the emotional turmoil of the characters involved.}
\\ \bottomrule
\end{tabular}
\caption{The screenplay summary for the \textit{Guiding Light} episode (aired 01-25-05). Our screenplay summary is able to recall 67\% of the visual information from the groundtruth summmary.} \label{tab:screenplay_summ_app}
\end{table*}

\begin{table*}[hbt]
\center
\small
\begin{tabular}{p{15cm}}
\begin{large}
    Gemini Summary of \textit{Guiding Light} episode (aired 01-25-05)
\end{large} \\
\toprule
\texttt{\textbf{1.} \hly{Reva and Josh are on a date}. Josh is trying to have a good time with Reva, but she is distracted, constantly worried about her son Jonathan and wanting to call home. It is revealed later that Jonathan is under the care of a nurse due to \hly{his being in a wheelchair}. Reva doesn’t trust the nurse with the medication, however, Josh tries to convince her that Jonathan is fine.}

\texttt{\textbf{2.} \hly{Cassie comes to Jonathan’s house, finds him alone and sitting in a wheelchair} \hlred{by the fireplace,} \hlgr{with some beer bottles} \hly{by his side.} Cassie wants to know what happened between her and Jonathan on the night before the wedding. \hly{She gives him a card that he sent her}; it read, “Stork doesn’t judge. A baby is a blessing even if it is conceived the night before the wedding.” Jonathan plays dumb, but eventually, Cassie reveals that Jonathan drugged her and she doesn’t remember what happened, and whether they slept together or not.}

\texttt{\textbf{3.} Gus is talking to Alan. Gus wants to find Beth because he thinks that she might know something about Phillip’s death, and he wants to clear Harley’s name. Alan, however, believes that Beth doesn’t know anything. He wants Gus removed from the company because ever since Phillip’s death, Gus has not been a good Spaulding. Gus remembers that he lied for Alan and Alexandra regarding the drugs, which cost him his wife. As far as Alan is concerned, Gus is siding with Harley and this will cost Beth her kids. Gus reminds Alan that Harley is also a mother and her son is Alan’s grandson.}

\texttt{\textbf{4.} Harley is talking to Beth. Beth admits to Harley that Phillip was going to take her children away and she had to stop him. She was going to ask Phillip to take her and the kids away. Harley thinks that was a smart idea to make Phillip reveal where the kids were. Beth confesses that she was going to leave with Phillip. Harley wants to know how she could rob another mother of her kids. Beth says she should ask Harley, and what did she do when she was planning to leave? Harley says she doesn’t remember if she killed Phillip. Beth says that the only weapon she had was history and she knew that Phillip would not believe her that night, so she changed her mind about leaving.}

\texttt{\textbf{5.} \hlgr{Gus meets with Lizzie in the house} and he asks her where her mother is. He explains he needs to talk to her, ask her a few questions about the night Phillip was killed. \hly{Lizzie doesn’t like Gus} and thinks that he doesn’t believe Harley is innocent. Gus claims that the evidence against Harley is circumstantial. Lizzie still refuses to tell Gus where her mother is.}

\texttt{\textbf{6.} \hly{Reva goes home and finds Cassie there.} \hlgr{She is angry.} Cassie is accused of wanting to kill Jonathan, because Jonathan is drunk. Cassie and Josh reveal to Reva what Jonathan did to Cassie on the night before the wedding. Reva finds out that Jonathan drugged and slept with Cassie. \hly{Reva and Josh have an argument} over Jonathan’s behavior, and about how Reva always takes Jonathan’s side.}

\texttt{\textbf{7.} Gus questions Lizzie about the night her father died and where she and her mother were. Lizzie claims that she was with her mother the entire time and that she didn’t kill Phillip. Gus suspects that Lizzie is protecting her mother by giving her an alibi.}

\texttt{\textbf{8.} Harley tells Beth to keep their conversation about the night Phillip died a secret between them. Beth owes Harley one. Gus calls Harley asking if she found Beth. Harley says that Beth was innocent. Gus thinks they can find another lead now, but Harley says she cannot do it anymore. During their conversation, \hly{Harley takes something} \hlred{out of her handbag and puts it back. She drops her handbag and something falls out, but she doesn’t pick it up.} \hly{She seems worried and scared.}}
\\ \bottomrule
\end{tabular}
\caption{Gemini's generated summary for the \textit{Guiding Light} episode (aired 01-25-05). Gemini's summary is able to recall 17\% of the visual information from the groundtruth summary.} \label{tab:gemini_end_to_end}
\end{table*}

\begin{table*}[hbt]
\center
\small
\begin{tabular}{p{15cm}}
\begin{large}
    VLog Summary of \textit{Guiding Light} episode (aired 01-25-05)
\end{large} \\
\toprule
\texttt{Based on the dialogue and visual details provided in the video, there are several subplots that can be identified:}

\texttt{\textbf{1.} The first subplot involves~\hly{a man in a gray jacket and white shirt} being questioned about the night his father died. He is asked about his mother's whereabouts and what she was wearing that night. The man insists his mother is innocent and refuses to provide further information, leading to suspicion from the questioner.}

\texttt{\textbf{2.} The second subplot features~\hly{a woman in a white robe and another woman} discussing a shared secret. They agree not to judge each other for their actions that night and decide to keep the information between themselves.}

\texttt{\textbf{3.} The third subplot revolves around \hlred{a man} \hly{talking on a cell phone while} \hlred{wearing a jacket}. He informs someone named Gus that a woman named Beth is innocent and did not see Phillip on the night he was killed. The man expresses reluctance to point fingers at other suspects and seems to be struggling with his own actions.}

\texttt{\textbf{4.} The fourth subplot shows~\hly{a woman} \hlred{holding a leaf up to her face} and expressing a desire for justice. She threatens someone named Phillip and vows to make him pay for his actions.}

\texttt{\textbf{5.} The fifth subplot involves~\hly{a woman in a blue shirt} \hlred{holding a remote control} and interacting with a nurse. She expresses concern about her son's medication and decides to leave him in the nurse's care while \hly{she steps away for a moment.}}

\texttt{\textbf{6.} The final subplot features~\hly{a man} \hlred{in a black jacket talking on a cell phone} and confronting someone named Alan. The man is determined to clear Harley's name and warns Alan not to interfere with his efforts.}

\texttt{These subplots involve a mix of mystery, secrets, suspicions, and confrontations, adding depth to the overall narrative of the video.}
\\ \bottomrule
\end{tabular}
\caption{VLog's generated summary for the \textit{Guiding Light} episode (aired 01-25-05). VLog's summary is able to recall 0\% of the visual information from the groundtruth summary.}
\end{table*}

\begin{table*}[hbt]
\center
\small
\begin{tabular}{p{15cm}}
\begin{large}
    Modular-Kosmos Summary of \textit{Guiding Light} episode (aired 01-25-05)
\end{large} \\
\toprule
\texttt{
\hlred{At the hospital}, Reva finds out that Jonathan has had his full dose of her fertility drugs and that he has no memory of the night of the wedding. \hlgr{At Company}, Alan tells Gus that he needs to remove Gus from the company. Gus tells Alan that he is trying to clear Harley's name. \hlred{Lizzie goes to the police station} and tells them that Harley did not kill her father. Gus tries to convince her that Harley is innocent. Alan tells Alan to get rid of Gus. Alan asks Gus if he knows something about the night Phillip died. Gus says that he knows nothing. Alan says that Alan needs to have Gus removed from Company. \hlred{At the farm}, Cassie asks Jonathan if he remembers the night before the wedding, but he says he doesn't remember. She asks him if they did the deed. He tells her that he does, but she says that she doesn't have to tell him. She tells him that she is on fertility drugs. He Reva tells Cassie that she's been drinking all night long and taking painkillers. Reva asks Cassie to give her the pills. Cassie tells Reva that the night before her wedding she was staying at Jonathan's place. Josh tells Jonathan that Reva threw him down the stairs. Jonathan tells Josh that he doesn't know what's going on. Lizzie tells Gus that Phillip was shot dead. Gus tells her that she needs to tell him what happened the night Phillip died. Harley tells Beth that she doesn't want her children to grow up without Phillip. Beth tells Harley that she didn't kill Phillip. Harley is upset and tells Beth to tell her what happened to Phillip. Gus asks her if she wants to talk to him. Harley says she's tired and wants to get out of the car. \hlred{Gus takes her to the police station}, where she tells him that Phillip is dead. Harley asks if she's going to tell Gus what happened. Gus
}\\ \bottomrule
\end{tabular}
\caption{Modular-Kosmos's generated summary for the \textit{Guiding Light} episode (aired 01-25-05). Modular-Kosmos's summary is able to recall 17\% of the visual information from the groundtruth summary.}
\end{table*}

\twocolumn

\end{document}